\documentclass[letterpaper, 10 pt, conference]{ieeeconf}  
\usepackage{amsmath}
\usepackage{graphicx}
\usepackage{multicol}
\usepackage{subcaption}
\usepackage{siunitx}
\usepackage{colortbl}
\usepackage{xcolor}
\usepackage{balance}
\usepackage{booktabs}
\usepackage{float}
\usepackage{titlesec}
\usepackage{tabularx}
\usepackage{placeins}

\titlespacing*{\subsubsection}{1pt}{*1}{*1}

\IEEEoverridecommandlockouts                              

\usepackage{graphics} 
\usepackage{mathptmx} 
\usepackage{times} 
\usepackage{amsmath} 
\usepackage{amssymb}  

\title{\LARGE \bf
YawSitter: Modeling and Controlling a Tail-Sitter UAV with Enhanced Yaw Control  
}

\author{Amir Habel, Fawad Mehboob, Jeffrin Sam,  Clement Fortin, Dzmitry Tsetserukou%
\thanks{The authors are with the Intelligent Space Robotics Laboratory, Skolkovo Institute of Science and Technology, Bolshoy Boulevard 30, bld. 1, 121205, Moscow, Russia
\tt \{Amir.Habel, Fawad.Mehboob, Jeffrin.Sam,  C.Fortin, D.Tsetserukou\}@skoltech.ru}} 

\begin{document}

\renewcommand{\thesubsection}{\Alph{subsection}}
\renewcommand{\thesubsubsection}{\thesubsection.\Roman{subsubsection}}

\makeatletter
\renewcommand{\@seccntformat}[1]{%
  \ifnum\pdfstrcmp{#1}{subsubsection}=0
    \thesubsubsection\ %
  \else
    \csname the#1\endcsname\ %
  \fi
}
\makeatother

\maketitle
\thispagestyle{empty}
\pagestyle{empty}

\begin{abstract}



Achieving precise lateral motion modeling and decoupled control in hover remains a significant challenge for tail-sitter Unmanned Aerial Vehicles (UAVs), primarily due to complex aerodynamic couplings and the absence of well-defined lateral dynamics. This paper presents a novel modeling and control strategy that enhances yaw authority and lateral motion by introducing a sideslip force model derived from differential propeller slipstream effects acting on the fuselage under differential thrust. The resulting lateral force along the body $y$-axis enables yaw-based lateral position control without inducing roll coupling.

The control framework employs a YXZ Euler rotation formulation to accurately represent attitude and incorporate gravitational components while directly controlling yaw in the $y$-axis, thereby improving lateral dynamic behavior and avoiding singularities.
The proposed approach is validated through trajectory-tracking simulations conducted in a Unity-based environment. Tests on both rectangular and circular paths in hover mode demonstrate stable performance, with low mean absolute position errors and yaw deviations constrained within \SI{5.688}{\degree}. These results confirm the effectiveness of the proposed lateral force generation model and provide a foundation for the development of agile, hover-capable tail-sitter UAVs.

\end{abstract}

\section{Introduction}
Tail-sitter Vertical Take-off and Landing (VTOL) fixed-wing Unmanned Aerial Vehicles (UAVs) combine the endurance of fixed-wing aircraft with vertical takeoff and landing capabilities, eliminating the need for a runway. Their simple design consists of a flying wing with large control surfaces and thrusters, but no vertical tail. This design enables transitions from vertical hover to level flight, where lift shifts from thrust to the fixed-wing, making them ideal for applications like surveillance, inspection, and payload delivery in confined or remote areas. Tail-sitters are categorized into mono-thrust transitioning (MTT), differential thrust transitioning (DTT), and collective thrust transitioning (CTT) types, as detailed in \cite{c1}. This study focuses on the CTT type, which uses coupled thrust and control surface actions.

Tail-sitter UAVs exhibit nonlinear and time-varying dynamics, particularly during hover-to-level transitions, posing significant modeling and control challenges \cite{c9}. The exact 6-DOF rigid body dynamics and aerodynamic effects are difficult to formulate, especially in the lateral direction. A rigorous dynamical model is found in \cite{c3}, with a detailed analysis of aerodynamic effects presented in \cite{c2} using a component breakdown approach. 

\begin{figure}[H]
    \centering
    \includegraphics[width=1\linewidth]{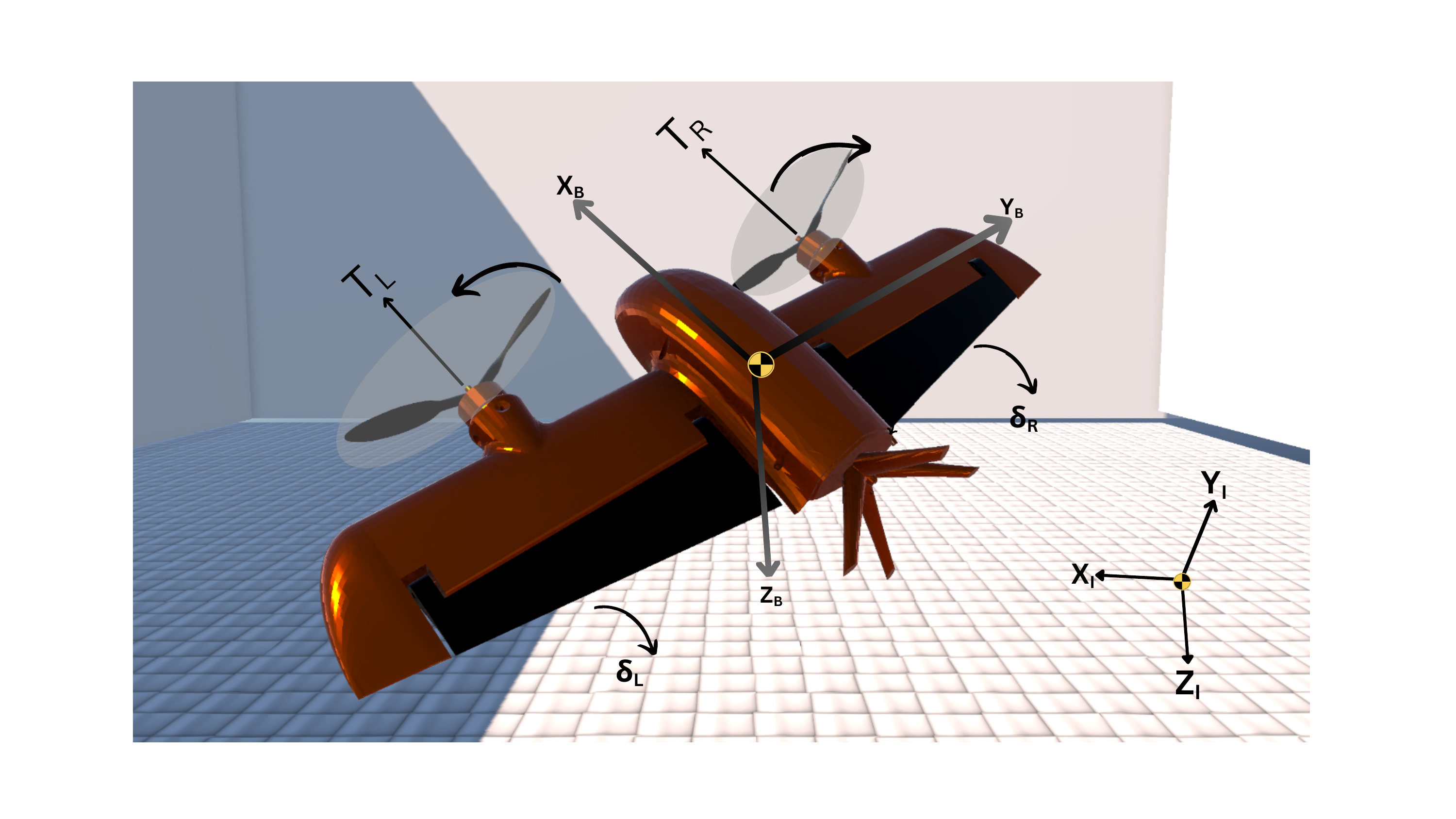}
    \caption{Tail-sitter frames description in UNITY simulation}
    \label{fig:Fig1_Frame_Diagram}
\end{figure}

Similarly, in \cite{c17}, the dynamics are rigorously defined, and the problem of lateral motion control is mentioned but not fully addressed. Some studies are also focused on using data-driven approaches addressing nonlinear dynamics in post stall regions \cite{c4}, but lateral motion control is missing. Moreover, the dynamical model are often further simplified while implementing control \cite{c3}.

Effective control of these underactuated systems is critical. Nonlinear quaternion-based cascaded PID controllers, comprising position, attitude, and thrust components, are common, as seen in \cite{c5, c7, c3}. Hierarchical PID with mode-switching \cite{c8} and smooth attitude control \cite{c6} address transition challenges. However, poor yaw authority and absent y-direction forces limit sideways motion and yaw control. Recent studies propose solutions like L1 neural network-based adaptive control \cite{c13}, optimal transitioning frameworks \cite{c9}, Dynamic Front and Back Transition Corridors \cite{c14}, successive linearization MPC \cite{c15}, unified cascaded PID frameworks \cite{c16}, and incremental nonlinear dynamic inversion \cite{c17}. Moreover, in \cite{c18}, Lyapunov-based control is used for 2D hovering to address uncertainties like gusts.

Despite these advances, inadequate yaw authority remains a limitation. This study presents a dual-thruster, no-stabilizer CTT tail-sitter UAV, enhancing yaw authority in both dynamics and control while retaining a cascaded PID framework, addressing the prior shortcomings.

\section{Methodology}

The system development methodology follows the workflow that uses aerodynamic analysis for selecting the airfoil (MH150) and deriving parameters ($C_L, C_D$, flap effects) via CFD, presented in Fig. (\ref{fig:aero_coeffs}-$b$), and Fig. (\ref{fig:aero_coeffs}-$c$). A 3D model was designed (SolidWorks), supporting physical fabrication for further real implementation with the model parameters as represented in Table \ref{tab:aircraft_parameters_combined}.


We developed a non-linear dynamical model that incorporates a novel lateral force modeling, generated by a differential slipstream of the propeller, to provide lateral force and improve yaw authority. The YXZ rotation matrix convention was utilized for accurate attitude representation and gravity compensation. A PID/PD-based controller was designed to leverage these model enhancements for precise hover trajectory tracking.

Validation was performed using a hybrid Python-Unity simulation framework, with Python computing control actions using the proposed dynamical model and Unity providing visualization. Rectangular and circular trajectories were tested in hover to evaluate yaw authority and lateral control. The methodology supports building a physical prototype and integrating controllers for real-time development via a proposed simulation framework using UNITY. This platform facilitates advancing future research for Tail-sitter UAVs.

\section{Dynamical Model} 

The system is modeled as a rigid body tail-sitter UAV with two fixed motors and two control surfaces (flaps), which generate thrust forces ($T_L, T_R$) and deflections ($\delta_L, \delta_R$). The system dynamics are governed by Newton's equations of motion in the body-fixed and inertial frames:
\begin{equation}
\label{dynamics}
\begin{split}
\dot{P}_I &= R^b_i (\phi,\theta, \psi) V_B \\
\dot{V}_B &= \frac{F_B}{m} - \omega_B \times V_B \\
\dot{\eta} &= W(\phi, \theta) \omega_B \\
\dot{\omega}_B &= J^{-1} (M_B - \omega_B \times J\omega_B)
\end{split}
\end{equation}
Where ${\dot{P}}_I$, ${\dot{V}}_B$ are the positions and velocity derivative in the inertial and body frames, respectively, Fig. \ref{fig:Fig1_Frame_Diagram}. ${R}(\psi, \theta, \phi)$ is the rotation matrix transforming the velocities from the body frame to the inertial frame using Euler angles $(\phi, \theta, \psi)$. ${V}_B = [u, v, w]^T$ and ${\omega}_B = [P, Q, R]^T$ are the translational and rotational velocities, respectively, 
defined in the body frame, and $\times$ denotes the 3D cross product. $\dot{{\eta}}$ represents the time rate of change in Euler angles $(Roll, Pitch, Yaw)$, related to the body angular velocities by the matrix ${W}(\phi, \theta)$. ${J}$ is the vehicle's inertia matrix, assumed diagonal $(J_{xx}, J_{yy}, J_{zz})$ and derived from the 3D model. ${F}_B$ and ${M}_B$ are the total translational force and rotational moment, respectively, acting on the vehicle's Center of Mass (CM) in the body frame.

The dynamics are nonlinear, with control inputs entering via ${F}_B$ and ${M}_B$. Detailed expansion of these terms, relating forces and moments to control inputs ($T_L, T_R, \delta_L, \delta_R$), is provided in subsequent sections.

\subsection{Forces and Moments Modeling}
\subsubsection{Transitional Forces}

The term $F_B$ is the sum of gravity $(F_g)$, thrust $(F_{thrust})$, aerodynamic $(F_{aero})$, and flap $(F_{flap})$ forces as shown in the Equ~\eqref{force}.
\begin{equation}
\label{force}
F_{total} = F_{gravity} + F_{thrust} + F_{aero} + F_{flaps}
\end{equation}

 where $F_{thrust}$ is defined as:
 \begin{equation}
 \label{thrust}
 F_{thrust} = \begin{bmatrix} T_L + T_R \\ F_{Y_B} \\ 0 \end{bmatrix}
 \end{equation}

The left and right motor thrusts are denoted by $T_L$ and $T_R$, respectively. The term $F_{Y_B}$ represents the lateral force arising from differential thrust, caused by the asymmetric dynamic pressure between the propeller slipstream and the fuselage sides. This effect is incorporated into the model to account for the side-force generated by unequal thrust levels.

The induced velocity of the propeller, simplified from the momentum theory, is given as follows:
\begin{equation}
V_{s,i} = \sqrt{\frac{4T_i}{\rho \pi R_s^2}}
\end{equation}
while the dynamic flow pressure can be described as: 
\begin{equation}
q_{i} = \frac{1}{2} \rho V_{s, i}^2 = \frac{2T_i}{\pi R_s^2}
\end{equation}
Therefore, the difference between the dynamic pressure of each propeller can be described as: 

\begin{equation}
q_L - q_R  = \frac{2(T_L - T_R)}{\pi R_s^2}
\end{equation}
Finally, we can define the side force due to the differential thrust as follows:
\begin{equation}
F_{y_B} = (q_L - q_R) S_{f} C_Y = K_f \Delta T
\end{equation}
Here $C_Y$ is a coefficient that has been tuned for more stable results and $K_f = \frac{2 S_{f} C_Y}{\pi R_s^2}$

$F_g$ is the representation of the gravity force in the body frame after transformation to the body frame:
\begin{equation}
\label{equ:4_gravity}
F_g = R^i_b\begin{bmatrix} 0 \\ 0 \\ mg   \end{bmatrix} 
\end{equation}

The aerodynamic forces are defined as:
\begin{equation}
\label{equ:equ5_Aero_Forces}
F_{Aero} = \begin{bmatrix} L \sin(\alpha) - D\cos(\alpha) 
\\ 0 
\\ -L \cos(\alpha) - D\sin(\alpha) \end{bmatrix} \end{equation}

$L$, and $D$ are the lift and drag forces given by the following relation:
\begin{equation}
\label{equ:equ6_Areodyynamic_forces}
\begin{split}
L = Pd \ S_{Wing} \ C_L \\
D = Pd \ S_{Wing} \ C_D \\
\end{split}
\end{equation}
Here, $Pd$ is the dynamic pressure over the tail-sitter defined as:

\begin{equation}
\label{equ7dynamicpressure}
\begin{split}
Pd= \frac{1}{2} \rho V_s^2 
\end{split}
\end{equation}

Where $S_{wing}$ is the wing area, $\rho$ is the air density at sea level,  $C_L$ and $C_D$ are mean aerodynamic coefficients obtained from the 3D CFD simulation described in Section \ref{Section_Aerodynamic_Coefficient}. 

Finally, the force contribution by the flap is given by:

\begin{equation}
\label{flaps}
F_{flap} = \begin{bmatrix} 0\\0\\ Pd \ S_{f} \ C_{l_{\delta}}(\delta_l, \delta_r) \end{bmatrix}
\end{equation}

Here, $C_{l_{\delta}}$ is the lift coefficient due to flap deflections, $\delta_l$ and $\delta_r$ are left and right flap deflections in radians, and $S_{f}$ is the flap surface area.


The slipstream velocity $V_s$ is derived using momentum theory \cite{c10}
\begin{align}
T &= \rho \pi R_s^2 V_s (V_s - u) \label{thrust_slipstream} \\
V_s &= \frac{-\lVert u \rVert + \sqrt{\lVert u \rVert^2 + \frac{4T}{\rho \pi R_s^2}}}{2} \label{equ11_Slipstream_velocity}
\end{align}
The thrust $T$ is estimated based on the previous thrust command, $||u||$ is the scalar longitudinal velocity in the x-body axis, and $R_s$ is the slipstream radius (simplified as $0.707 R$, where $R$ is the propeller radius). This velocity is constrained to remain above the slipstream velocity at hover ($u = 0$, $T_{min} \approx mg$), approximately, preventing aggressive control when the slipstream velocity is low.

\subsubsection{Rotational Dynamics}

The total moment $M_B = [L, M, N]^T$ at the vehicle's CM includes contributions from thrust $(M_{thrust})$, aerodynamic $(M_{aero})$, and flap $(M_{flap})$ moments.

\begin{equation}
\label{moments}
\begin{split}
M_B = M_{Thrust} + M_{Aero} + M_{flap} 
\end{split}
\end{equation}

The moment generated by each motor due to thrust and propeller drag can be expressed as
\begin{equation}
\label{eq:m_thrust}
\mathbf{M}_{\text{thrust},i} =
\begin{bmatrix}
 \sum_{i=1}^{2} Q_i
 \\[4pt] 0 \\[4pt]  l_y \  \sum_{i=1}^{2} T_i
\end{bmatrix},
\end{equation}
where $T_i$ is the thrust produced by the motor $i$, $Q_i$ is the reaction torque of the propeller around the thrust axis, and $l_y$ is the lateral distance from the motor to the center of mass of the vehicle. The thrust generated by each motor is modeled as:
\begin{equation}
\label{eq:thrust_force}
T_i = (-1)^i C_T \, \omega_i^2,
\end{equation}
where $C_T$ is the thrust coefficient and $\omega_i$ is the motor angular speed. The corresponding reaction torque due to aerodynamic drag on the propeller blades is given by:
\begin{equation}
\label{eq:reaction_torque}
Q_i = -(-1)^i C_\mu \, \omega_i^2,
\end{equation}
where $C_\mu$ is the torque coefficient, and the factor $-(-1)^i$ accounts for the opposite spin directions of the left and right motors ($i= L \ , R$). The thrust components contribute to the total lift and yawing moment through differential thrust, while the reaction torques generate a rolling moment that depends on the difference in motor speeds.

Equation~\ref{equ21_m_aero} describes the aerodynamic pitching moment about the body $Y_B$-axis. This moment consists of the natural pitching moment and the additional moment produced by the aerodynamic force acting along the $Z_B$-axis at the center of mass offset. 

\begin{equation}
\label{equ21_m_aero}
M_{\text{aero}} =
\begin{bmatrix}
0 \\
M_{\text{pitch}} + F_{\text{Aero},z}(X_a - X_g) \\
0
\end{bmatrix}.
\end{equation}

The pitching moment component $M_{\text{pitch}}$ is defined as:

\begin{equation}
\label{Pitch_and_yaw_moments}
M_{\text{pitch}} = P_{d,\text{wing}}\,\bar{c}\,C_M,
\end{equation}

where $\bar{c}$ is the mean aerodynamic chord, $C_M$ is the pitching moment coefficient obtained from the aerodynamic analysis (Section~\ref{Section_Aerodynamic_Coefficient}), and $X_a$ and $X_g$ denote the aerodynamic center and the center of gravity, respectively.

The flap contribution is given by:
\begin{equation}
\label{m_flap}
M_{flap} = P_{d_{flap}} S_{f} \begin{bmatrix} b \ (C_{L_{L-f}}, C_{L_{R-f}})
\\
\bar{c} \ C_{M_{\delta}}(\delta_L, \delta_R)
\\
b \ (C_{L_{L-f}}, C_{L_{R-f}})\end{bmatrix}
\end{equation}

Where $C_{L_{L-f}}$ and $ C_{L_{R-f}}$ represent the dependence of the lift coefficients on the left and right flaps, respectively, and $C_{M_{\delta}}$ describes the moment coefficient due to unidirectional flap deflection.

\subsection{Aerodynamic Coefficient}
\label{Section_Aerodynamic_Coefficient}
The UAV operates in a wide flight envelope with varying aerodynamics coefficients such as the lift coefficient ($C_l$), drag coefficient ($C_d$), and moment coefficient ($C_m$), which must be modeled using flight data due to their dependence on the angle of attack $\alpha$ and flap deflections $\delta_l$ and $\delta_r$. A key parameter in modeling is the propeller slipstream velocity (Equ. \ref{equ11_Slipstream_velocity}). 

The relationship between these parameters is as follows:
\begin{equation}
\label{CL}
C_L = f(C_{L_{\alpha}},\ C_{L_{\delta}}\ (\delta_L, \delta_R),\ C_o)
\end{equation}
$C_L$ is a function of the $C_{L_{\alpha}}$ defined according to the change of the $C_L$ vs $\alpha$, and the flap deflections scaled by the slope of $C_l$-$\alpha$ curve due to the flap deflection, which in the linear domain is defined in \cite{c11}:

\begin{equation}
\label{CL_alpha}
C_{L_{\alpha}} = \frac{2\pi cos\Lambda}{\frac{2\pi cos\Lambda}{A_R} + \sqrt{1+(\frac{2\pi cos\Lambda}{A_R})^2}}
\end{equation}

$\Lambda$ here is the sweep angle and $A_R$ is the aspect ratio which can be defined as $A_R = b^2 / S_{wing}$.
Similarly, we model the drag coefficient as:

\begin{equation}
\label{CD}
C_D = C_{D0} + \frac{C_L^2}{\pi e_o A_R}
\end{equation}

The zero-lift drag coefficient, \( C_{D_0} \), accounts for skin friction drag, while the Oswald efficiency factor, \( e \), typically ranges from 0.8 to 0.95, as noted in \cite{c12}. The pitching moment coefficient, \( C_M \), is expressed as a function of the moment due to angle of attack, control surface deflections, and the zero-angle moment, given by:
\begin{equation}
\label{eq:CM}
C_M = f(C_{M_{\alpha}}, C_{M_{\delta}} (\delta_L, \delta_R), C_{M_0}),
\end{equation}
where \( C_{M_{\alpha}} \) is the pitching moment slope, \( C_{M_{\delta}} \) is the moment due to left and right control surface deflections (\( \delta_L \), \( \delta_R \)), and \( C_{M_0} \) is the zero-angle pitching moment.

As the angle of attack $\alpha$ increases, the wing approaches stall, leading to a reduction in lift. The aerodynamic model incorporates two flow-separation mechanisms, Progressive boundary-layer separation and Turbulent leading-edge vortices. In conventional forward flight, the instantaneous $\alpha$ is calculated using body-frame velocity components:
\begin{equation}
\label{eq:alpha_forward}
\alpha = \arctan\left(\frac{w}{u}\right),
\end{equation}
where \( u \) and \( w \) represent the longitudinal and vertical body velocities, respectively. During hovering, where \( u \approx 0 \), the classical $\alpha$ definition becomes ill-posed. To address this, the longitudinal velocity is augmented by the propeller slipstream velocity, \( V_{\text{prop}} \), yielding:
\begin{equation}
\label{eq:alpha_hover}
\alpha = \arctan\left(\frac{w}{u + V_{\text{prop}}}\right).
\end{equation}

This formulation is valid across the flight envelope, accounting for propeller-induced inflow in low-forward-velocity regimes.


In Fig. (\ref{fig:aero_coeffs}-$a$), the curves show the dependence of the pitching moment coefficient for 3 cases, i.e., maximum flap deflection, minimum flap deflection, and no flap deflection.
Similarly, we modeled the dependence of the lift and drag coefficients as shown in Fig. (\ref{fig:aero_coeffs}-$b$) and Fig. (\ref{fig:aero_coeffs}-$c$):



\begin{figure}[!htbp]
    \centering

    \begin{subfigure}[h]{0.2\textwidth}
        \includegraphics[width=\textwidth]{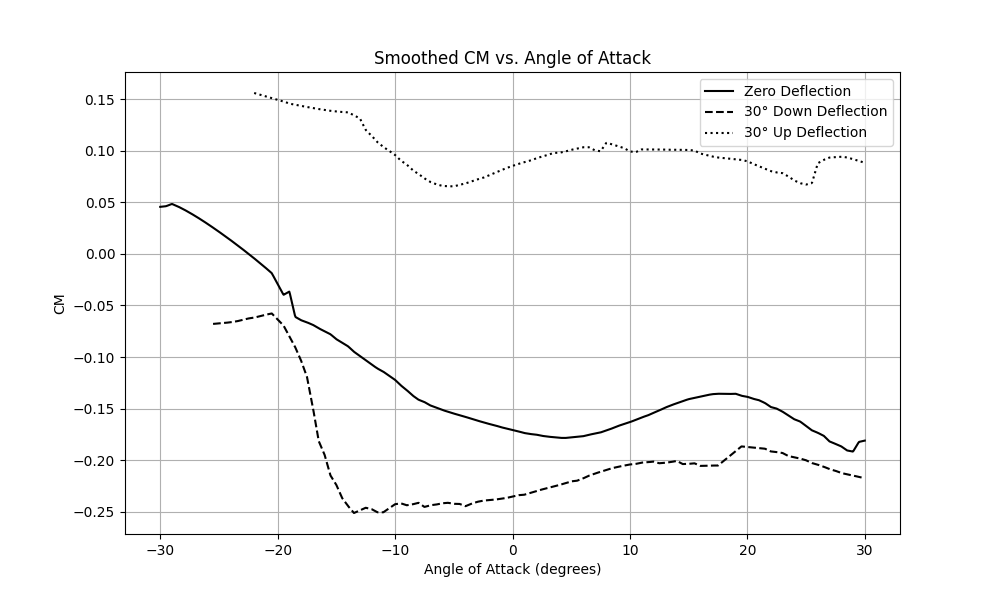}
        \caption{}
    \end{subfigure}
    \begin{subfigure}[h]{0.2\textwidth}
        \includegraphics[width=\textwidth]{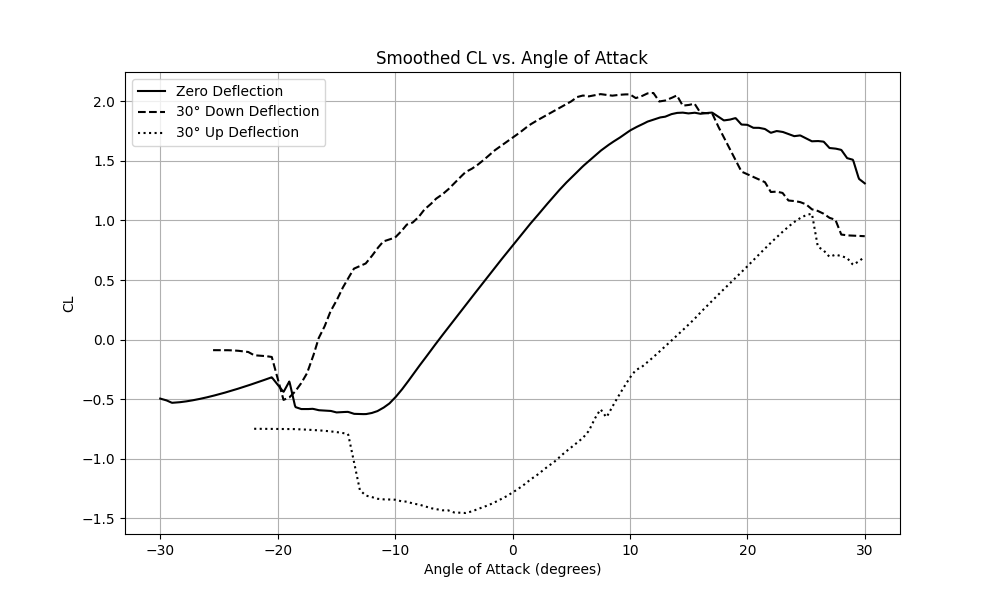}
        \caption{}
    \end{subfigure}
    \begin{subfigure}[h]{0.2\textwidth}
        \includegraphics[width=\textwidth]{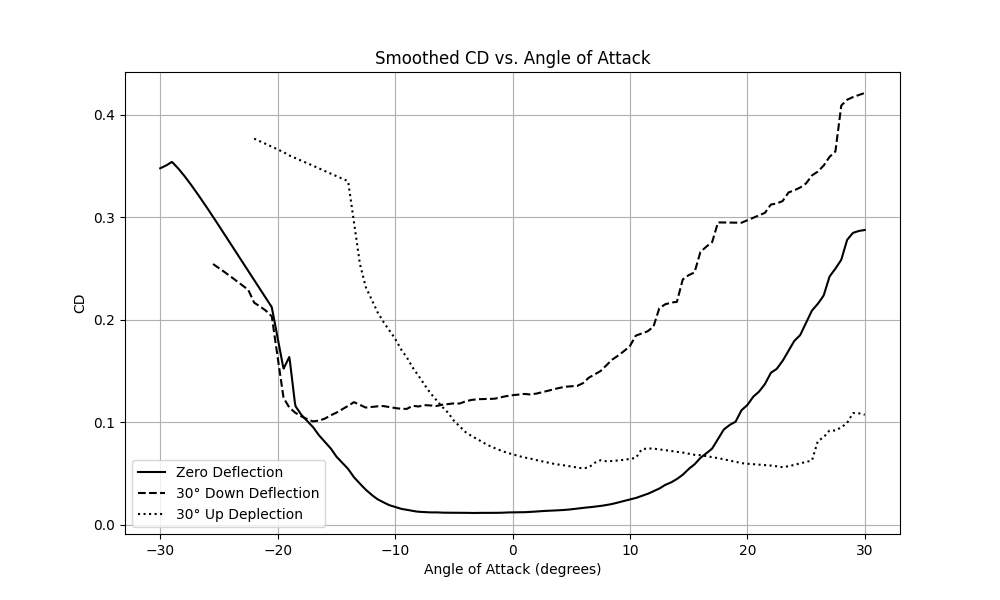}
        \caption{}
    \end{subfigure}

    \caption{Aerodynamic coefficient variation with angle of attack and flap deflection: (a) Moment coefficient \( C_m \), (b) Lift coefficient \( C_L \), and (c) Drag coefficient \( C_D \).}
    \label{fig:aero_coeffs}
\end{figure}


\section{Nonlinear PID Controller Design}
The control system presented here is designed to track a reference position and attitude of a vehicle, such as an unmanned aerial vehicle (UAV) or underwater vehicle, by computing desired velocities, forces, and moments. The position controller generates inertial-frame velocity commands using a PID strategy based on position error. These velocities are compared to the vehicle's inertial velocity, obtained by transforming body-frame velocities using a rotation matrix. A velocity PID controller then computes desired inertial forces, adjusted for gravity, which are transformed to the body frame for actuator commands. The attitude controller commands body angular rates to achieve desired Euler angles (roll, pitch, yaw), and a moment controller generates body-frame moments. Finally, an inner loop maps these forces and moments to control inputs (thrust and flap deflections) using sum and difference variables. This study provides detailed derivations, explanations, and considerations for practical implementation, including handling Euler angle singularities.


\begin{figure*}[!htbp]
    \centering
    \includegraphics[width=1\linewidth]{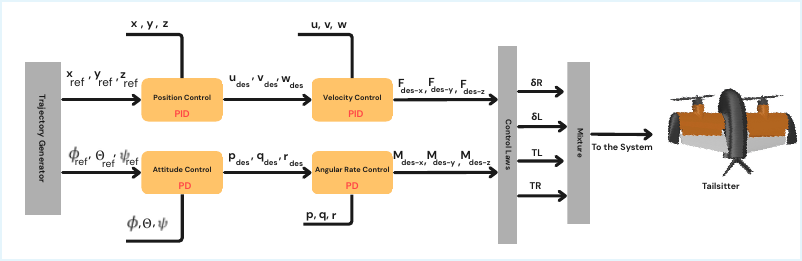}
    \caption{Controller diagram logic. The trajectory generator provides reference 
    positions and attitudes to the position and attitude controllers. 
    The position PID outputs desired linear velocities, while the attitude PD outputs 
    desired angular rates. Velocity and rate controllers compute the required forces 
    and moments, which are mapped by control laws to actuator commands 
    $(\delta_R, \delta_L, T_L, T_R)$.}
    \label{fig:Controller_Diagram}
\end{figure*}

\subsection{Position Controller}
\label{Position_Controller}
\noindent
The position controller computes the desired velocity command in the inertial frame to track a reference position: 
\begin{equation}
{X}_{ref} = [x_{ref}, y_{ref}, z_{ref}]^T
\end{equation}
It uses a PID (Proportional-Integral-Derivative) control strategy based on the position error 
\begin{equation}
{e}_p = {X}_{ref} - {X}
\end{equation}
where \({X} = [x, y, z]^T\) is the current position.

\subsection{Velocity Controller}
\label{Velocity_Controller}
The desired velocity command in the inertial frame, \(\mathbf{V}_{DCI} = [\dot{x}_{des}, \dot{y}_{des}, \dot{z}_{des}]^T\), is given by:

\begin{equation}
\mathbf{V}_{DCI} = \mathbf{K}_{pp} (e_p) 
\\ + \mathbf{K}_{PI} \int (e_p) \, dt 
\\ + \mathbf{K}_{PD} \frac{d}{dt} (e_p)
\end{equation}


Here, $\mathbf{K}_{PP}$, $\mathbf{K}_{PI}$, and $\mathbf{K}_{PD}$ denote the diagonal matrices of proportional, integral, and derivative gains for the position errors, respectively. The subscripts \textit{DCI} and following \textit{DCB} represent the desired commanded parameters expressed in the inertial and body frames, respectively. The current inertial velocity, $\mathbf{V}_I = [\dot{x}, \dot{y}, \dot{z}]^T$, is obtained by transforming the body-frame velocities, $\mathbf{V}_B = [u, v, w]^T$, using the rotation matrix from the body frame to the inertial frame, $R^b_i$.

\begin{equation}
\mathbf{V}_I = R^b_i \mathbf{V}_{B},
\end{equation}

where $R^b_i$ is the rotation matrix for a 3-2-1 Euler angle sequence (yaw \(\psi\), pitch \(\theta\), roll \(\phi\)).

\subsection{Desired Forces Calculations}
\label{Desired_Forces}
The desired forces in the inertial frame, \(\mathbf{F}_{DCI} = [F_{x,des}, F_{y,des}, F_{z,des}]^T\), are computed using a PID controller on the velocity error ${e}_v = {V}_{DCI} - {V}$, scaled by the vehicle's mass \(m\):

\begin{equation}
\mathbf{F}_{\text{DCI}} = m \Bigg[ \mathbf{K}_{pv} (e_v) + \mathbf{K}_{vi} \int (e_v) \, dt + \mathbf{K}_{vD} \frac{d}{dt} (e_v) \Bigg]
\end{equation}
where, the \(\mathbf{K}_{vp}\), \(\mathbf{K}_{vi}\), and \(\mathbf{K}_{vD}\) are the diagonal matrices of the proportional, integral, and derivative gains for the velocities of $X, Y, Z$ directions. 
While, to generate actuator commands, the desired inertial forces are transformed to the body frame using the inverse rotation matrix \(R^i_b = (R^b_i)^{-1} = (R^b_i)^T\):

\begin{equation}
\mathbf{F}_{DCB} = R^i_b \mathbf{F}_{DCI}
\end{equation}

where \(\mathbf{F}_{DCB} = [F_{des,x-body}, F_{des,y-body}, F_{des,z-body}]^T\). Since \(R^b_i\) is orthogonal, \(R^i_b = (R^b_i)^T\).

\subsection{Attitude Control}
\label{Attitude_Controller}

The attitude controller commands body angular rates to achieve desired Euler angles (\(\phi_{ref}, \theta_{ref}, \psi_{ref}\)) relative to the current angles (\(\phi, \theta, \psi\)) in the 3-2-1 rotation sequence (yaw, pitch, roll). A PD (Proportional-Derivative) control strategy is used to compute the desired angular rates, improving tracking performance over a proportional-only approach by eliminating steady-state errors and damping oscillations. Discontinuities in attitude errors are handled to ensure smooth control, and units of the gains are analyzed to maintain physical consistency.

\subsection{Desired Angular Rates}
The desired body angular rates \(\boldsymbol{\omega}_{des} = [P_{des}, Q_{des}, R_{des}]^T\), representing roll, pitch, and yaw rates, are computed using a PD controller on the attitude errors \(e_\phi = \phi_{ref} - \phi\), \(e_\theta = \theta_{ref} - \theta\), and \(e_\psi = \psi_{ref} - \psi\):

\begin{align}
P_{des} &= K_{p\phi} e_\phi + K_{D\phi} \frac{d e_\phi}{dt}, \\
Q_{des} &= K_{p\theta} e_\theta + K_{D\theta} \frac{d e_\theta}{dt}, \\
R_{des} &= K_{p\psi} e_\psi +  K_{D\psi} \frac{d e_\psi}{dt},
\end{align}


For handling the discontinuity in the attitude errors due to the periodicity of the angles, we normalized the angular errors.






\subsubsection{Desired Moments}
The desired moments can be calculated according to the desired angular rates on the body frame as the following: 
\begin{equation}
    M_{des-Body} = J^{-1} [K_{\omega p}(e_\omega) + K_{\omega I} \int(e_\omega)dt + K_{\omega D} \frac{d(e_\omega )}{dt}  ]
\end{equation}
Where the $J^{-1}$ is the diagonal moment of inertia of the vehicle, and the $e_\omega$ is the error of the angular rates of the tail-sitter $e_\omega = \omega_{des} - \omega$.

\subsection{Control Allocation Using Sum and Difference Variables}

To allocate the desired forces and moments to the actuators, a sum-and-difference formulation is adopted. 
We define the total and differential thrusts as
\begin{equation}
\label{sum_and_difference_Thrust_formulation}
T = T_L + T_R, 
\qquad 
\Delta T = T_L - T_R
\end{equation}
and the symmetric and differential flap deflections as
\begin{equation}
\label{sum_and_difference_Deflection_formulation}
\delta_s = \frac{\delta_L + \delta_R}{2}, 
\qquad 
\Delta \delta = \delta_L - \delta_R
\end{equation}

\subsubsection{Intermediate Control Quantities}

The desired differential thrust required to produce the commanded yawing moment is
\begin{equation}
\Delta T = \frac{M_{\text{des},z,\text{Body}}}{L_y} + F_{\text{des},y,\text{Body}} ,
\label{eq:deltaT}
\end{equation}
where $L_y$ is the lateral motor separation.

The required differential flap deflection is obtained from the commanded rolling moment, including the coupling due to propeller reaction torque:
\begin{equation}
\Delta \delta = 
\frac{
M_{\text{des},x,\text{Body}} + c_\mu \dfrac{M_{\text{des},z,\text{Body}}}{L_y}
}{
q_{\text{flap}} \, b \, C_{L_\delta}},
\label{eq:deltadelta}
\end{equation}
where $c_\mu$ is the reaction torque coefficient, $q_{\text{flap}}$ is the dynamic pressure at the flaps, $b$ is the wing span, and $C_{L_\delta}$ is the flap lift coefficient.

The symmetric flap deflection required to generate the necessary vertical force component is
\begin{equation}
\delta_s =
\frac{
F_{\text{des},z,\text{Body}} 
- F_{g,z,\text{Body}} 
- F_{\text{Aero},z,\text{Body}}
}{
q_{\text{flap}} \, C_{L_\delta}},
\label{eq:deltas}
\end{equation}
where $F_{g,z,\text{Body}}$ and $F_{\text{Aero},z,\text{Body}}$ are the gravitational and aerodynamic force components along the $z$-axis.

The total thrust required from the motors is then computed as
\begin{equation}
T = 
F_{\text{des},x,\text{Body}} 
- F_{g,x,\text{Body}} 
- F_{\text{Aero},x,\text{Body}} 
- q_{\text{flap}} C_{L_\delta} \delta_s,
\label{eq:totalT}
\end{equation}
where the final term accounts for the contribution of symmetric flap deflection to the net $x$-axis force.




\subsubsection{Final Actuator Commands Matrix}

The motor thrusts and flap deflections can be compactly expressed using a sum-and-difference mapping:

\begin{equation}
\begin{bmatrix}
T_L \\[4pt]
T_R
\end{bmatrix}
=
\frac{1}{2}
\begin{bmatrix}
1 & 1 \\[4pt]
1 & -1
\end{bmatrix}
\begin{bmatrix}
T \\[4pt]
\Delta T
\end{bmatrix},
\label{eq:thrust_matrix}
\end{equation}

\begin{equation}
\begin{bmatrix}
\delta_L \\[4pt]
\delta_R
\end{bmatrix}
=
\frac{1}{2}
\begin{bmatrix}
1 & 1 \\[4pt]
1 & -1
\end{bmatrix}
\begin{bmatrix}
\delta_s \\[4pt]
\Delta \delta
\end{bmatrix}.
\label{eq:elevon_matrix}
\end{equation}


The above formulation provides a direct mapping from the desired body-frame forces and moments $(F_{\text{des}}, M_{\text{des}})$ to actuator commands $(T_L, T_R, \delta_L, \delta_R)$. 
By including the motor reaction torque term in Eq.~\eqref{eq:deltadelta} and subtracting the flap-induced $x$-force in Eq.~\eqref{eq:totalT}, this control allocation accounts for roll–yaw coupling and the aerodynamic contribution of the flaps. 
The resulting structure is compact and suitable for real-time implementation, while remaining consistent with the physical actuator effects of the tailsitter UAV.

\begin{figure*}[!t]
 \centering
\begin{subfigure}[h]{0.3\textwidth}
\includegraphics[width=\textwidth]{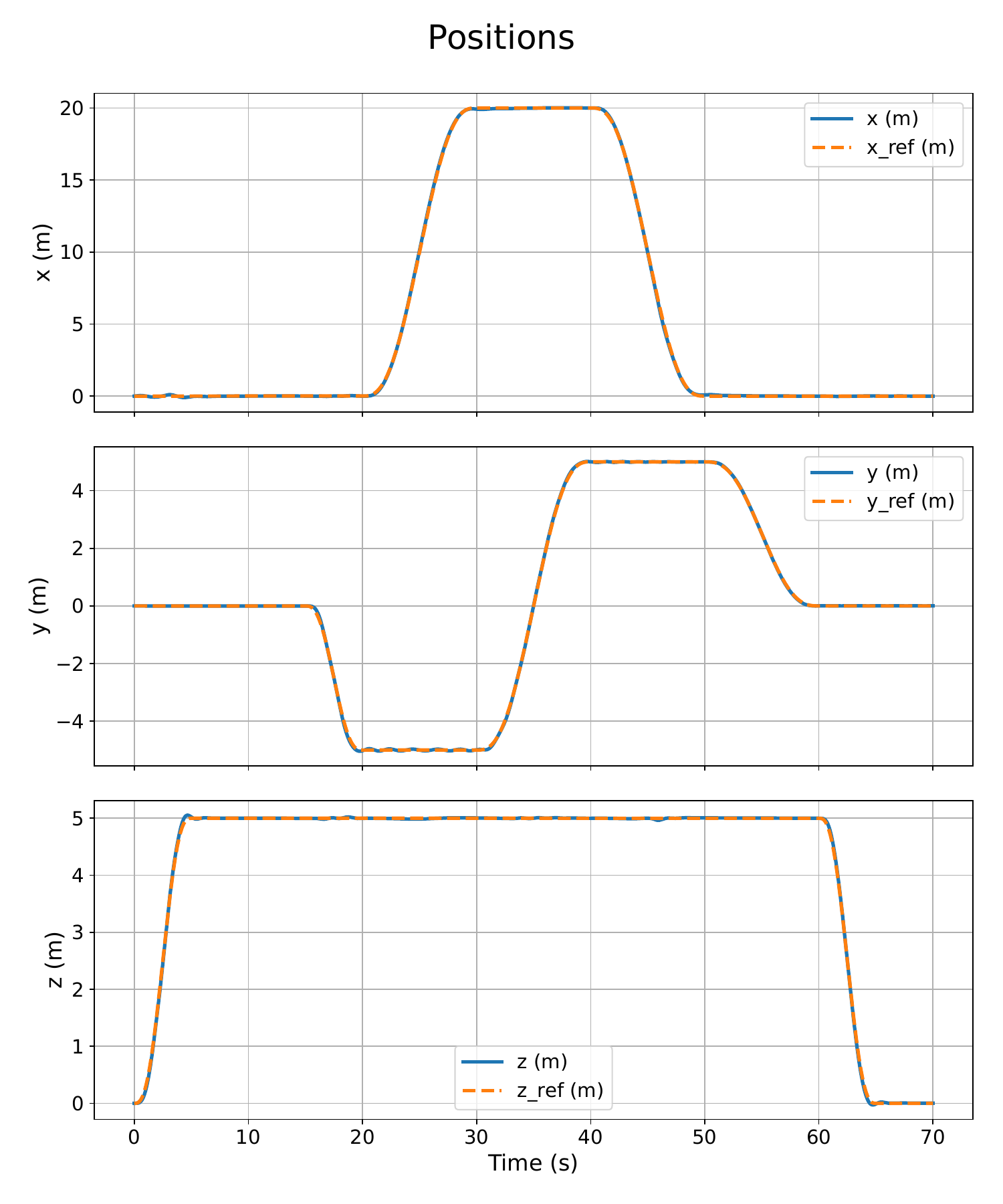}
\caption{}
\end{subfigure}
\begin{subfigure}[h]{0.3\textwidth}
\includegraphics[width=\textwidth]{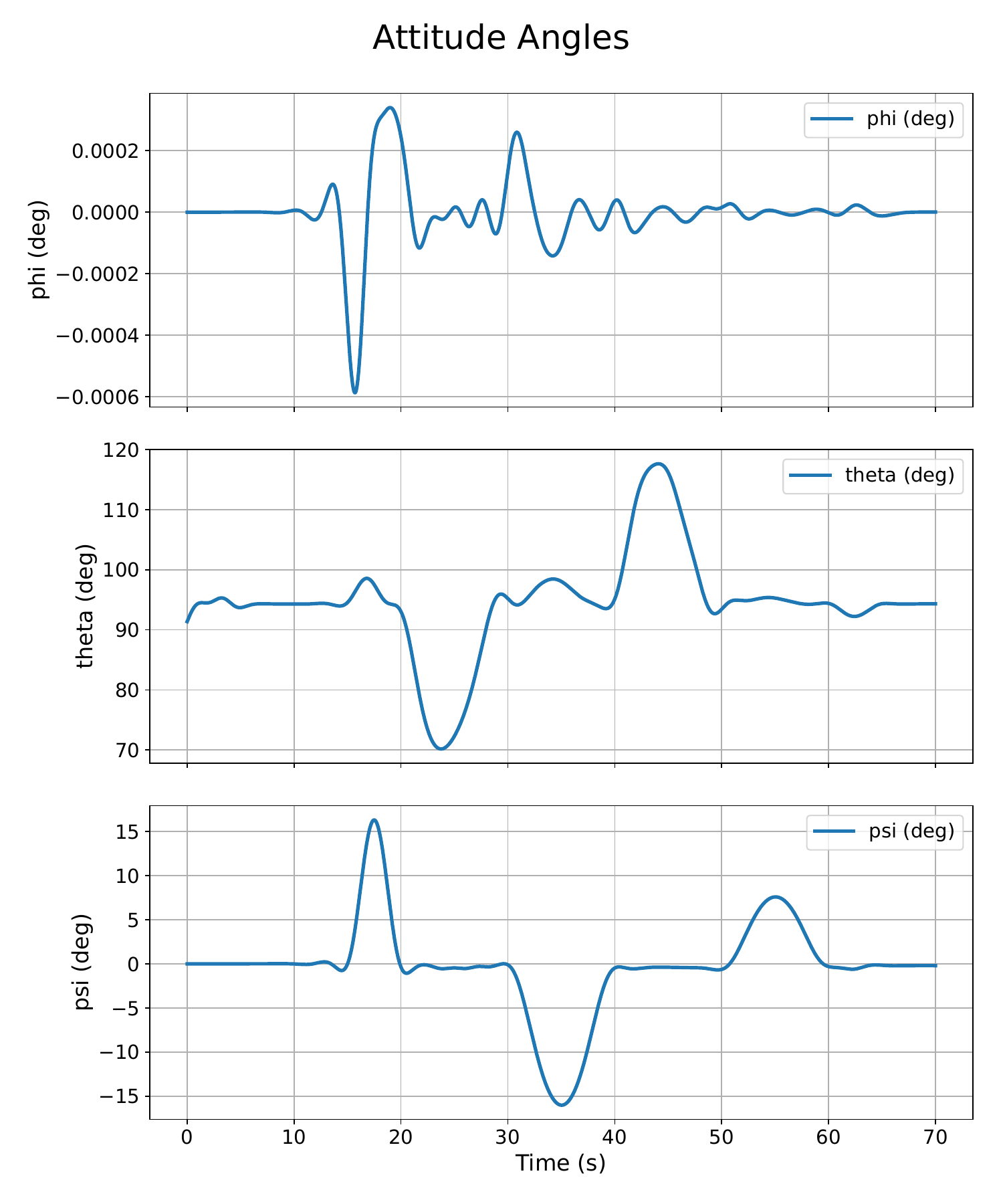}
\caption{}
\end{subfigure}
\begin{subfigure}[h]{0.26\textwidth}
\includegraphics[width=\textwidth]{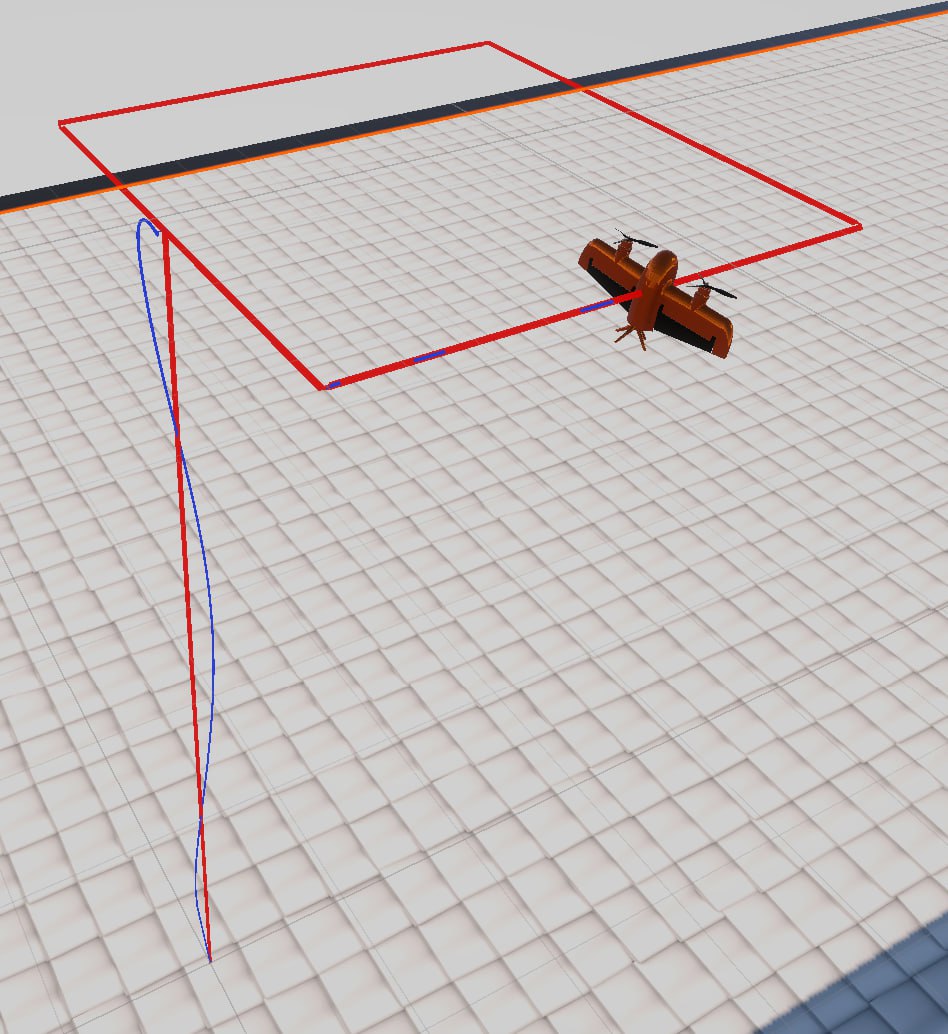}
\caption{}
\end{subfigure}
 \caption{Rectangular trajectory tracking of tail-sitter,(a) Positions responses(m) for $(x, y, z)$, (b)Attitude response  of $(\phi, \theta, \psi)$ $^\circ$, and (c) Real-time simulation in UNITY.}
\label{fig:Rectangular_Results}
 \end{figure*}

\begin{figure*}
 \centering
    \begin{subfigure}[h]{0.3\textwidth}
        \includegraphics[width=\textwidth]{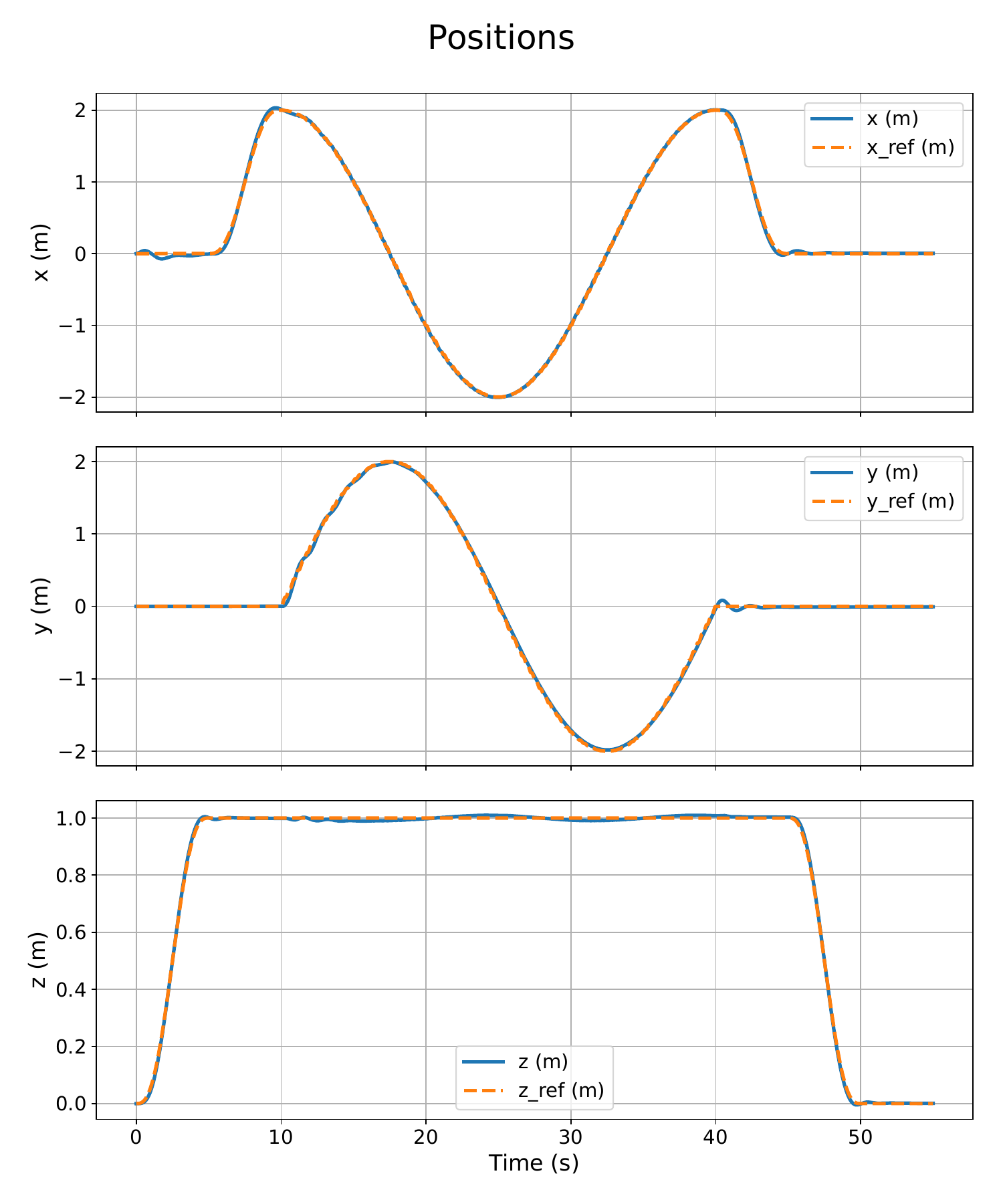}
        \caption{}
    \end{subfigure}
    \begin{subfigure}[h]{0.3\textwidth}
        \includegraphics[width=\textwidth]{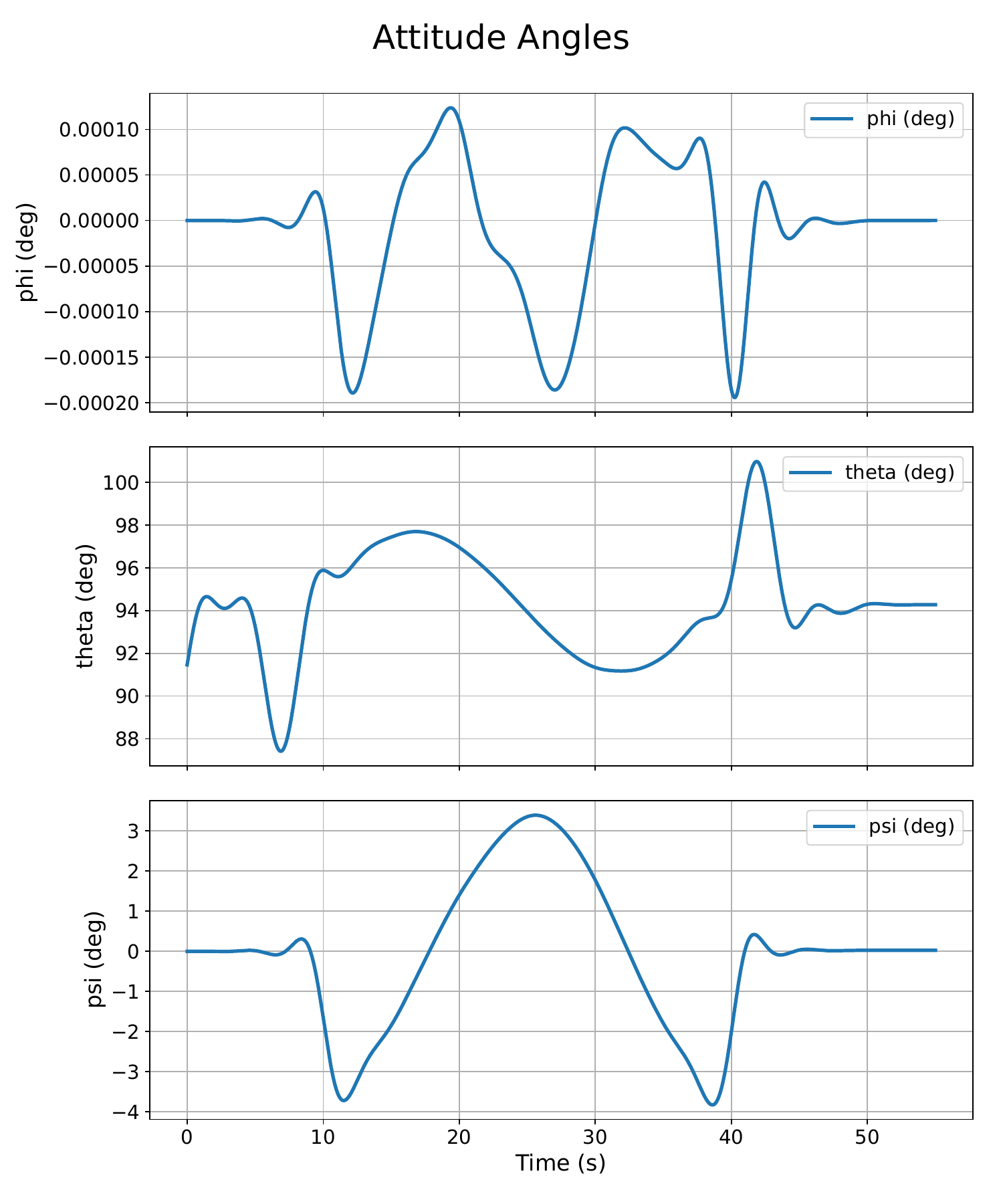}
        \caption{}
    \end{subfigure}
    \begin{subfigure}[h]{0.3\textwidth}
        \includegraphics[width=\textwidth]{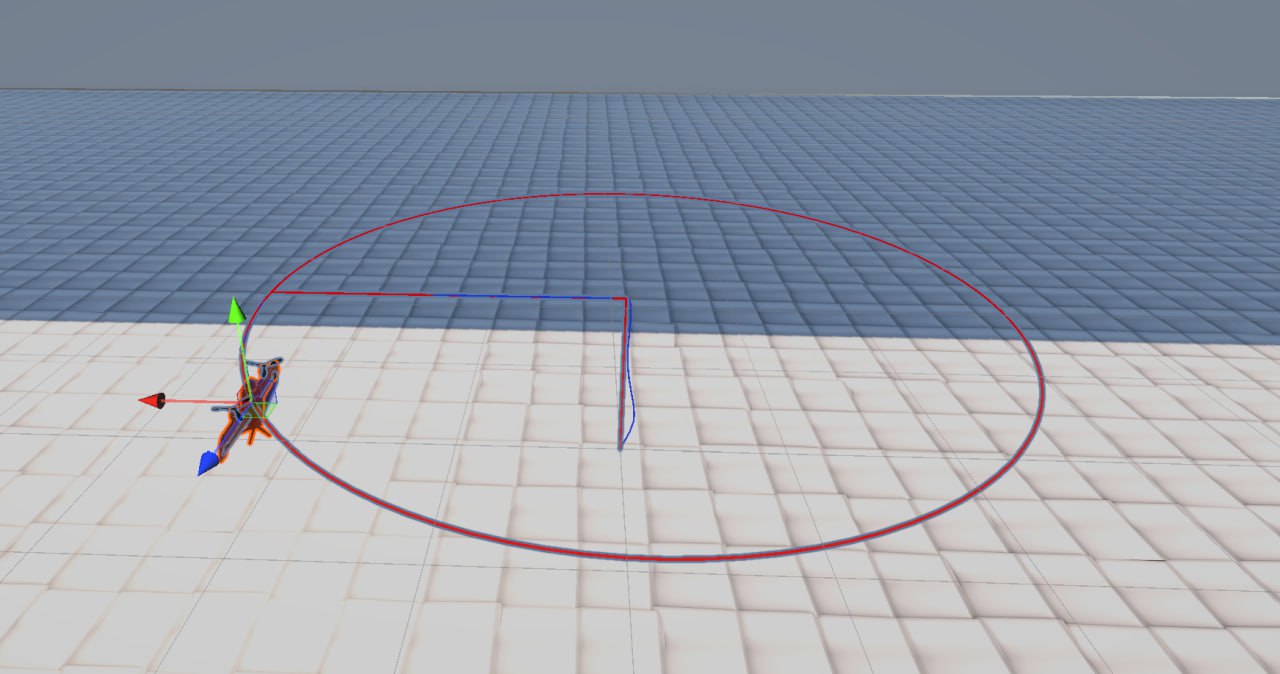}
        \caption{}
 \end{subfigure}
 \caption{Circular trajectory tracking of tail-sitter,(a) Positions responses for $(x, y, z)$, (b)Attitude response of $(\phi, \theta, \psi)$ $^\circ$, and (c) Real-time simulation in UNITY.}
 \label{fig:Circulare_Results}
\end{figure*}

\section{Results}
\label{sec:Results}


This section presents simulation results validating the proposed lateral control strategy for a tail-sitter UAV in hover mode. The model employs differential thrust, derived from asymmetric propeller slipstream effects, to generate lateral forces that improve yaw authority and $y$-axis position tracking while avoiding roll coupling. Additionally, the use of a YXZ Euler rotation convention incorporates gravitational terms that further enhance the lateral dynamics. 

To comprehensively evaluate tracking performance, five key metrics are considered: the mean absolute error (MAE), the root mean square error (RMSE), the integrated absolute error (IAE), the maximum overshoot and the final error at the last waypoint.  

The simulation framework, developed in Python and Unity 6, assesses the controller performance on both rectangular and circular trajectories with respect to waypoint tracking, yaw regulation, and altitude control.

\subsection{Rectangular Trajectory}





A rectangular trajectory of \( \SI{20}{\meter} \times \SI{5}{\meter} \) was simulated over \( \SI{70}{\second} \) to evaluate lateral motion and yaw control in hover. 
The UAV demonstrated accurate position and altitude tracking, as summarized in Table~\ref{tab:rectangular_tracking_errors}, which reports the errors at key waypoints and the overall performance metrics. 
The table shows low mean absolute errors, small final position errors, and limited overshoot along all axes, indicating precise trajectory tracking. 

Yaw deviations peaked at approximately \( -1.60^\circ \) but were corrected effectively, while the hover pitch angle remained close to \( 93^\circ \). 
The total thrust varied between \( \SI{10.835}{\newton} \) and \( \SI{14.815}{\newton} \), supporting stable hover and transition phases. 
The results confirm effective yaw correction, precise tracking, and stable hover performance (Fig.~\ref{fig:Rectangular_Results}).

\begin{table}[H]
    \centering
    \caption{Position Tracking Errors at Key Waypoints and Overall Performance Metrics (Rectangular Trajectory)}
    \label{tab:rectangular_tracking_errors}
    \renewcommand{\arraystretch}{1}
    \setlength{\tabcolsep}{4pt}
    \begin{tabular}{c@{\hspace{5pt}}c@{\hspace{5pt}}c@{\hspace{5pt}}c@{\hspace{5pt}}c}
        \toprule
        \textbf{Time (s)} & \textbf{Waypoint ($x, y, z$) (m)} & \textbf{Error $x$ (m)} & \textbf{Error $y$ (m)} & \textbf{Error $z$ (m)} \\
        \midrule
        20.0 & (0.00, -5.0, -5.0)  &  0.0052  &  0.0251  & -0.0013 \\
        30.0 & (20.0, -5.0, -5.0) &  0.0828  & -0.0159  &  0.0056 \\
        40.0 & (20.0, \ 5.0, -5.0) &  0.0005  & -0.0007  & -0.0002 \\
        50.0 & (0.00,  \ 5.0, -5.0) & -0.0863  & -0.0020  &  0.0060 \\
        60.0 & (0.00,  \ 0.0, -5.0) & -0.0016  & -0.0014  &  0.0001 \\
        65.0 & (0.00,  \ 0.00,  0.00) & -0.0129  & -0.0045  & -0.0112 \\
        \midrule
        \multicolumn{4}{c}  {\textbf{Overall Performance Metrics}} \\
        \midrule
        \textbf{Metric}          & \textbf{$x$ (m)} & \textbf{$y$ (m)} & \textbf{$z$ (m)} \\
        \midrule
        \textbf{MAE}             & 0.0317 & 0.0135 & 0.0099 \\
        \textbf{RMSE}            & 0.0501 & 0.0236 & 0.0187 \\
        \textbf{IAE (m$\cdot$s)} & 2.2225 & 0.9445 & 0.6901 \\
        \textbf{Max Overshoot}   & 0.2360 & 0.1183 & 0.0855 \\
        \textbf{Final Error}     & -0.0007 & -0.0023 & 0.0026 \\
        \bottomrule
    \end{tabular}
\end{table}


\subsection{Circular Trajectory}


A circular trajectory with a radius of \( \SI{2}{\meter} \) and an altitude of \( \SI{1}{\meter} \) was simulated for \( \SI{55}{\second} \) to evaluate the continuous tracking of the yaw and the lateral control. 
Position tracking performance, including errors at key waypoints and overall performance metrics, is summarized in Table~\ref{tab:circular_tracking_errors}. 
The table shows low mean absolute errors, small final position errors, and limited overshoot along all axes, confirming precise trajectory tracking. 

The roll remained stable at approximately \( \phi \pm 0.001^\circ \), while the pitch varied between \( 87.2^\circ \) and \( 101.3^\circ \). 
The yaw deviation peaked at \( 5.7^\circ \), but was effectively compensated by the proposed control strategy. 
The thrust output remained within \( \SIrange{12.526}{13.452}{\newton} \), and the flap deflections remained below \( \pm 21.1^\circ \), well within the actuator limits. 
These results demonstrate accurate path tracking, stable attitude regulation, and effective yaw control, as visualized in Fig.~\ref{fig:Circulare_Results}.

\begin{table}[H]
    \centering
    \caption{Position Tracking Errors at Key Waypoints and Overall Performance Metrics (Circular Trajectory)}
    \label{tab:circular_tracking_errors}
    \renewcommand{\arraystretch}{1.}
    \setlength{\tabcolsep}{4pt}
    \begin{tabular}{c c c c}
        \toprule
        \textbf{Time (s)} & \textbf{Error $x$ (m)} & \textbf{Error $y$ (m)} & \textbf{Error $z$ (m)} \\
        \midrule
        5.0    &  0.0057  &  0.0000  &  0.0012 \\
        15.0   & -0.0127  &  0.0177  & -0.0090 \\
        25.0   &  0.0055  & -0.0336  &  0.0094 \\
        35.0   &  0.0090  & -0.0014  & -0.0008 \\
        45.0   & -0.0007  &  0.0111  &  0.0030 \\
        55.0   & -0.0037  &  0.0056  &  0.0009 \\
        \midrule
        \multicolumn{4}{c}{\textbf{Overall Performance Metrics}} \\
        \midrule
        \textbf{Metric}          & \textbf{$x$ (m)} & \textbf{$y$ (m)} & \textbf{$z$ (m)} \\
        \midrule
        \textbf{MAE}             & 0.0178 & 0.0162 & 0.0056 \\
        \textbf{RMSE}            & 0.0238 & 0.0242 & 0.0066 \\
        \textbf{IAE (m$\cdot$s)} & 0.9788 & 0.8933 & 0.3061 \\
        \textbf{Max Overshoot}   & 0.0728 & 0.1330 & 0.0143 \\
        \textbf{Final Error}     & 0.0037 & 0.0056 & 0.0009 \\
        \bottomrule
    \end{tabular}
\end{table}

\section{Conclusion}

This work introduced and validated a modeling approach to account for lateral dynamics in tail-sitter UAVs during hover. By exploiting asymmetric propeller slipstream, differential thrust was used to generate lateral forces, enhancing yaw authority and enabling $y$-axis tracking without roll coupling. The use of a YXZ Euler rotation matrix further improved lateral dynamics by including gravity-coupled effects.

Simulations in a Python–Unity framework showed low mean absolute position errors in both rectangular and circular trajectories (Section~\ref{sec:Results}). Compared to conventional VTOL strategies that tilt the platform or redirect thrust, the proposed method achieves lateral control while preserving roll neutrality—beneficial for tasks requiring high stability, such as inspection, payload delivery, and perching.

Identified limitations include the need for careful roll tuning, pitch deviations due to the high-camber airfoil, and potential Euler singularities during aggressive maneuvers. Overall, results demonstrate the potential of slipstream-based lateral control to achieve decoupled and precise motion in hovering regimes.

\section{Future Work}


Future work will focus on overcoming identified limitations and expanding control capabilities. Attitude representation singularities will be eliminated by employing a YawSitter quaternion-based controller to provide greater robustness during high maneuvers. PID control will be enhanced for pitch/roll stability over airfoil challenges, as well as exploring nonlinear control approaches to enable agile maneuvers. 

These controllers are verified to involve extensive Software-in-Loop (SIL) testing before being physically implemented. These procedures are built atop the side force model's conceptual levels to result in more resilient and responsive tail-sitter control in a variety of flight regimes.



\section*{APPENDIX}

\begin{table}[H] 
\centering
\caption{Tail-sitter Parameters}
\label{tab:aircraft_parameters_combined}
\renewcommand{\arraystretch}{1.2} 
\begin{tabular}{llll} 
\toprule 
\multicolumn{2}{c}{\textbf{Symbol}} & \multicolumn{2}{c}{\textbf{Value}} \\ 
\midrule 

\multicolumn{4}{c}{\textbf{Model Parameters}} \\ 
\midrule
$m$ & $1.076$ \si{\kilogram} & $S_{\text{Wing}}$ & $0.0882$ \si{\meter\squared} \\
\midrule
$S_{\text{f}}$ & 0.02 \si{\meter\squared} & $S_{\text{F}}$ & $0.067$ \si{\meter\squared} \\
\midrule
$b$ & $0.63$ \si{\meter} & ${\bar{C}}$ & 0.14 \si{\meter} \\
\midrule
$X_g$ & $0.026$ \si{\meter} & $X_a$ & $0.03$ \si{\meter} \\
\midrule
AR & $\approx 4.5$  & $l_{\text{distance}}$ & $0.168$ \si{\meter} \\
\midrule
$J_{xx}$ & $0.0134$ kg \si{\meter\squared} & $J_{yy}$ & $0.015438$ kg \si{\meter\squared} \\
\midrule
$J_{zz}$ & $0.02462$ kg \si{\meter\squared} & $R_{\text{prop}}$ & $0.11425$ \si{\meter} \\
\midrule
$K_t$ & $0.0051$ \si{\per\meter} & ${K}_v$ & $1.24$ 1/s \\
\midrule
$e$ & $0.8$  & \multicolumn{2}{c}{} \\ 
\midrule 

\multicolumn{4}{c}{\textbf{Aerodynamic Parameters}} \\ 
\midrule
$C_{L_\alpha}$ & $4.4$ /rad & $C_{M_\alpha}$ & $-0.2$ /rad \\
\midrule
$C_{L_\delta}$ & $1.5$ /rad & $C_{M_\delta}$ & $-1.0$ /rad \\
\midrule
$C_{L0}$ & $-0.2477$  & $C_{D0}$ & $0.05$  \\
\midrule
$C_{M0}$ & $-0.0390$  \\ 
\midrule 

\multicolumn{4}{c}{\textbf{Constraints}} \\ 
\midrule
$T_{\text{max}}$ & $22.0$ \si{\newton} & $T_{\text{min}}$ & $0.5$ \si{\newton} \\
\midrule
$\delta_{\text{max}}$ & 35 \si{\degree} & $\delta_{\text{min}}$ & -35 \si{\degree} \\
\bottomrule 
\end{tabular}
\end{table}

\begin{table}[H]
\centering
\caption{The Tuned Gains in Tail-Sitter UAV Control Loops}
\label{tab:pid_gains_initial}
\renewcommand{\arraystretch}{1.2}
\begin{tabular}{
>{\centering\arraybackslash}p{2.3cm}|
S[table-format=2.2, table-column-width=1.7cm]|
S[table-format=2.2, table-column-width=1.7cm]|
S[table-format=2.2, table-column-width=1.7cm]}
\hline
\textbf{States} &
\multicolumn{1}{S[table-format=2.2, table-column-width=1.7cm]}{\textbf{$K_p \ (S^{-1})$}} &
\textbf{$K_i \ (S^{-2})$} &
\textbf{$K_d$} \\
\hline
$x$ (pos)       & 2.84  & 0.012 & 0.55 \\
$y$ (pos)       & 2.22  & 0.01  & 0.52 \\
$z$ (pos)       & 1.21  & 0.05  & 0.71 \\
\hline
$v_x$ (vel)     & 1.22  & 0.15 & 0.25 \\
$v_y$ (vel)     & 1.12  & 0.11 & 0.25 \\
$v_z$ (vel)     & 1.57  & 0.22 & 0.30 \\
\hline
$\phi$ (att)    & 4.24  & 0.00 & 0.80 \\
$\theta$ (att)  & 5.31  & 0.00 & 1.25 \\
$\psi$ (att)    & 3.32  & 0.00 & 0.55 \\
\hline
$P$ (rate)      & 1.55  & 0.00 & 0.15 \\
$Q$ (rate)      & 2.50  & 0.00 & 0.05 \\
$R$ (rate)      & 1.25  & 0.00 & 0.05 \\
\hline
\end{tabular}
\end{table}

\end{document}